\documentclass[10pt]{elsart}
\journal{Theoretical Computer Science}
\newif \iffinal \finalfalse

\usepackage{amsmath,amssymb,latexsym,bm}
\usepackage{algorithm}
\usepackage{algorithmic}
\usepackage{times}
\usepackage[mathscr]{eucal}

\newtheorem{theorem}{Theorem}[section]

\newtheorem{lemma}[theorem]{Lemma}
\newtheorem{observation}[theorem]{Observation}
\newtheorem{corollary}[theorem]{Corollary}

\theoremstyle{definition}
\newtheorem{definition}[theorem]{Definition}
\theoremstyle{remark}
\newtheorem{example}[theorem]{Example}

\newcommand{\ignore}[1]{}
\newcommand{\eps}{\epsilon}

\newcommand{\E}{{\bf E}}

\newcommand{\poly}{\mathrm{poly}}

\newcommand{\bits}{\{-1,1\}}
\newcommand{\bn}{\bits^n}
\newcommand{\bnz}{\{0,1\}^n}
\newcommand{\bbn}{[b]^n}

\newcommand\conju[1]{\overline{#1}}
\newcommand{\sign}{\mathrm{sign}}
\newcommand{\fousumelt}{\hat{f}(\alpha)}
\newcommand{\dist}{\mathscr{D}}
\newcommand{\pru}{\omega_b}
\newcommand{\ghs}{\mathsf{GHS}}
\newcommand{\hs}{\mathsf{HS}}
\newcommand{\dnf}{\mathsf{DNF}}
\newcommand{\ACZERO}{\mathsf{AC^0}}

\newcommand{\mq}{\mathsf{MEM}(f)}
\newcommand{\exo}{\mathsf{EX}(f,\mathscr{D})}
\newcommand{\algo}{\mathcal{A}}
\newcommand{\cc}{\mathfrak{C}}
\newcommand\innp[2]{\langle {#1},{#2} \rangle}

\newcommand{\littlesum}{\mathop{{\textstyle \sum}}}
\newcommand{\littleprod}{\mathop{{\textstyle \prod}}}

\newcommand{\fisafunc}{f \colon \bn \to \bits}
\newcommand{\fisgfunc}{f \colon \bbn \to \bits}

\makeatother \iffinal
\newcommand{\fnote}[1]{}
\newcommand{\onote}[1]{}
\newcommand{\snote}[1]{}
\newcommand{\remove}[1]{}
\else
    \newcommand{\fnote}[1]{\footnote{{\bf [[Jon: {#1}\bf ]] }}}
    \newcommand{\snote}[1]{\footnote{{\bf [[Rocco: {#1}\bf ]] }}}
    \newcommand{\onote}[1]{\footnote{{\bf [[Ryan: {#1}\bf ]]}}}
    \newcommand{\remove}[1]{\par $<<${\it removed part}$>>$}
    \newcommand{\old}[1]{}
\fi

\begin{document}
\begin{frontmatter}

\title{Learning Unions of $\omega(1)$-Dimensional Rectangles}

\author[Alp]{Alp At\i c{\i}}
\ead{atici@math.columbia.edu},
\author[Rocco]{Rocco~A. Servedio\thanksref{roccogrant}\corauthref{cor}}
\corauth[cor]{Corresponding author. Phone: (212) 939-7065,
Fax: (212) 666-0140.}
\thanks[roccogrant]{Supported in part by NSF award CCF-0347282 and NSF
award CCF-0523664.}
\ead{rocco@cs.columbia.edu}

\address[Alp]{Department of Mathematics,
Columbia University,\\ 2990 Broadway, Mail Code: 4406,
New York, NY 10027}
\address[Rocco]{Department of Computer Science,
Columbia University,\\ 1214 Amsterdam Avenue, Mail Code: 0401,
New York, NY 10027
}

\begin{abstract}
We consider the problem of learning unions of rectangles over
the domain $\bbn$, in the uniform distribution membership query
learning setting, where both $b$ and $n$ are ``large''.
We obtain poly$(n, \log b)$-time algorithms for the following
classes:
\begin{itemize}
        \item $\poly (n \log b)$-way \textsc{Majority} of $O(\frac{\log(n \log b)} {\log \log(n \log b)})$-dimensional rectangles.
        \item Union of $\poly(\log(n \log b))$ many $O(\frac{\log^2 (n \log b)} {(\log \log(n \log b) \log \log \log (n \log b))^2})$-dimensional
            rectangles.
        \item $\poly (n \log b)$-way \textsc{Majority} of
$\poly (n \log b)$-\textsc{Or} of disjoint $O(\frac{\log(n \log b)}
{\log \log(n \log b)})$ dimensional rectangles.
\end{itemize}
Our main algorithmic tool
is an extension of Jackson's boosting- and Fourier-based Harmonic Sieve
algorithm \cite{JACKSON} to
the domain $\bbn$, building on work of Akavia {\em et al.} \cite{AGS}.
Other ingredients used to obtain the results stated above are techniques
from exact learning \cite{BK}
and ideas from recent work on learning augmented $\ACZERO$ circuits
\cite{JKS} and on representing Boolean functions as thresholds of parities
\cite{KS}.
\end{abstract}
\end{frontmatter}

\section{Introduction}

\subsection{Motivation} \label{sec:motivation}
The learnability of Boolean valued functions defined over the domain
$$\bbn=\{0, 1, \ldots, b-1\}^n$$ has long elicited interest in
computational learning theory literature.
In particular, much research has been done on learning
various classes of ``unions of rectangles'' over $\bbn$ (see e.g.
\cite{BK,CH,CM,GGM,JACKSON,MW}), where a rectangle is a
conjunction of properties of the form ``the value of attribute $x_i$ lies
in the range $[\alpha_i,\beta_i]$''.
One motivation for studying these classes is that they
are a natural analogue of classes of $\dnf$ (Disjunctive Normal Form) formulae
over $\bnz$; for instance, it is easy to see that
in the case $b=2$ any union of $s$ rectangles is simply a $\dnf$ with
$s$ terms.

Since the description length of a point $x \in \bbn$ is $n \log b$ bits,
a natural goal in learning functions over $\bbn$ is to obtain
algorithms which run in time $\poly(n \log b)$.
Throughout the article we refer to such algorithms with
$\poly(n \log b)$ runtime as {\em efficient} algorithms.
In this article we give efficient algorithms which can learn
several interesting classes of unions of rectangles over $\bbn$
in the model of uniform distribution learning with membership queries.

\subsection{Previous results}
In a breakthrough result a decade ago, Jackson \cite{JACKSON} gave the Harmonic
Sieve ($\hs$) algorithm and proved that it can learn any $s$-term $\dnf$
formula over $n$ Boolean variables in $\poly(n,s)$ time.
In fact, Jackson showed that the algorithm can learn any $s$-way majority
of parities in $\poly(n,s)$ time; this is a richer set of functions
which includes all $s$-term $\dnf$ formulae.
The $\hs$ algorithm works by boosting a Fourier-based weak learning
algorithm, which is a modified version of an earlier algorithm due
to Kushilevitz and Mansour \cite{KM}.

In \cite{JACKSON} Jackson also described an extension of the $\hs$
algorithm to the domain $\bbn$.  His main result for $\bbn$ is
an algorithm that can learn any union of $s$ rectangles over $\bbn$
in $\poly(s^{b \log \log b}, n)$ time; note that this runtime
is $\poly(n,s)$ if and only if $b$ is $\Theta(1)$ (and the runtime
is clearly exponential in $b$ for any $s$).

There has also been substantial work on learning various classes
of unions of rectangles over $\bbn$ in the more demanding model of exact
learning from membership and equivalence queries.
Some of the subclasses of unions of rectangles which have been considered
in this setting are
\begin{description}
\item[The dimension of each rectangle is $O(1)$:]
Beimel and Kushilevitz established an algorithm
learning any union of $s$ $O(1)$-dimensional rectangles over $\bbn$
using equivalence queries only, in $\poly(n,s,\log b)$ time steps \cite{BK}.

\item[The number of rectangles is limited:]
In \cite{BK} an algorithm is also given which exactly learns any union
of $O(\log n)$ many rectangles
in $\poly(n, \log b)$ time using membership
and equivalence queries.  Earlier, Maass and Warmuth \cite{MW} gave
an algorithm which uses only equivalence queries and can learn
any union of $O(1)$ rectangles in $\poly(n,\log b)$ time.

\item[The rectangles are disjoint:]  If no
input $x \in \bbn$ belongs to more than one rectangle, then
\cite{BK} can learn a union of $s$ such rectangles in $\poly(n,s,\log b)$
time with membership and equivalence queries.
\end{description}

\subsection{Our techniques and results}
Because efficient learnability is established for unions of $O(\log n)$ arbitrary dimensional
rectangles by \cite{BK} in a more demanding model, we are interested in achieving positive
results when the number of rectangles is strictly larger. Therefore all the cases we study
involve at least $\poly(\log (n \log b))$ and sometimes as many as $\poly(n \log b)$ rectangles.

We start by describing a new variant of the Harmonic Sieve algorithm for
learning functions defined over $\bbn$; we call this new algorithm
the Generalized Harmonic Sieve, or $\ghs$.
The key difference between $\ghs$ and Jackson's algorithm for $\bbn$
is that whereas Jackson's algorithm used a weak learning algorithm
whose runtime is $\poly(b)$, the $\ghs$ algorithm uses a $\poly(\log b)$
time weak learning algorithm described in recent work of Akavia {\em et al.}
\cite{AGS}.

We then apply $\ghs$ to learn various classes of functions defined in
terms of ``$b$-literals'' (see Section~\ref{sec:prelim} for a precise
definition; roughly speaking a $b$-literal is like a 1-dimensional
rectangle).  We first show the following result:

\begin{theorem}\label{thmapp1}
The concept class of $s$-way \textsc{Majority} of $r$-way
\textsc{Parity} of $b$-literals where $s=\poly (n \log b)$,
$r=O(\frac{\log(n \log b)} {\log \log (n \log b)})$ is efficiently
learnable using $\ghs$.
\end{theorem}

Learning this class has immediate applications for our goal of
``learning unions of rectangles''; in particular, it follows that

\begin{theorem}\label{finthm1} The concept class of
    $s$-way \textsc{Majority} of $r$-dimensional rectangles
    where $s=$ $\poly (n \log b)$,
    $r=O(\frac{\log(n \log b)} {\log \log(n \log b)})$ is efficiently
learnable using $\ghs$.
\end{theorem}

This clearly implies efficient learnability for unions (as opposed to
majorities) of $s$ such rectangles as well.

We then employ a technique of restricting the domain $\bbn$ to a much smaller
set and adaptively expanding this set as required. This approach was used in the
exact learning framework by Beimel and Kushilevitz \cite{BK}; by an appropriate
modification we adapt the underlying idea to the uniform distribution membership
query framework. Using this approach in conjunction with $\ghs$ we obtain almost a quadratic improvement in
the dimension of the rectangles if the number of terms is guaranteed to be small:

\begin{theorem} \label{finthm2}
The concept class of unions of $\poly(\log(n \log b))$ many
$r$-dimensional rectangles where $r=O(\frac{\log^2 (n \log b)}
{(\log \log(n \log b) \log \log \log (n \log b))^2})$ is efficiently
learnable via Algorithm $2$ (see Section~\ref{sec:grid}).
\end{theorem}

Finally we consider the case of disjoint rectangles (also studied by
\cite{BK} as mentioned above), and improve the depth of our circuits by $1$
provided that the rectangles connected to the same \textsc{Or} gate
are disjoint:

\begin{corollary}\label{fincor2} The concept class of
$s$-way \textsc{Majority} of $t$-way \textsc{Or} of
    disjoint $r$-dimensional rectangles where
    $s, t =\poly (n \log b)$,
    $r=O(\frac{\log(n \log b)} {\log \log(n \log b)})$ is efficiently learnable under $\ghs$.
\end{corollary}

\subsection{Organization} In Section 3 we describe the
Generalized Harmonic Sieve algorithm $\ghs$ which will be our main
tool for learning unions of rectangles. In Section $4$ we show that
$s$-way \textsc{Majority} of $r$-way \textsc{Parity} of $b$-literals
is efficiently learnable using $\ghs$ for suitable $r,s$; this
concept class turns out to be quite useful for learning unions of
rectangles. In Section $5$ we improve over the results of Section
$4$ slightly if the number of terms is small, by adaptively
selecting a small subset of $[b]$ in each dimension which is
sufficient for learning, and invoke $\ghs$ over the restricted
domain. In Section $6$ we explore the consequences of the results in
Sections $4$ and $5$ for the ultimate goal of learning unions of
rectangles.

\section{Preliminaries} \label{sec:prelim}

\subsection{The learning model}\label{sec:lrnmodel}
We are interested in Boolean functions defined over the
domain $\bbn$, where $[b]=\{0,1,\ldots,b-1\}$.
We view Boolean functions as mappings into $\{-1,1\}$ where $-1$
is associated with \textsc{True} and $1$ with \textsc{False}.

A \emph{concept class} $\cc$ is a collection of classes (sets) of
Boolean functions $\{C_{n,b} \colon n>0, b>1\}$ such that if $f\in
C_{n,b}$ then $\fisgfunc$.  As a simple example, consider the case
where $b=2$ and $\cc$ is the class of all monotone Boolean
conjunctions; then for each $n$ we have that $C_{n,b}$ is the set of
all Boolean conjunctions over a subset of the Boolean input
variables $x_1,\dots,x_n.$  Throughout this article we view both $n$
and $b$ as asymptotic parameters, and our goal, as mentioned in
Section~\ref{sec:motivation}, is to construct algorithms that learn
various classes $C_{n,b}$ in poly$(n, \log b)$ time. (Note that
given this goal, it only makes sense to attempt to learn concept
classes such that each concept in the class has ``description
length'' at most poly$(n \log b)$ bits. It will be clear that this
is the case for all the concept classes we consider.) We now
describe the uniform distribution membership query learning model
that we will consider.

A \emph{membership oracle} $\mq$ is an oracle which, when queried with input $x$, outputs the label $f(x)$ assigned by the target $f$ to
the input. Let $f\in C_{n,b}$ be an unknown member of the concept class and let $\algo$ be a randomized learning algorithm which takes as
input accuracy and confidence parameters $\epsilon, \delta$ and can invoke $\mq$. We say that \emph{$\algo$ learns $\cc$ under the
uniform distribution on $\bbn$} provided that given any $0< \epsilon,
\delta<1$ and access to $\mq$, with probability at least $1-\delta$
$\algo$ outputs an $\epsilon$-approximating hypothesis $h \colon \bbn \to \bits$
(which need not belong to $\cc$)
such that $\Pr_{x\in \bbn}[f(x)=h(x)]\geq 1-\epsilon$.

We are interested in computationally efficient learning algorithms.
We say that $\algo$ {\em learns $\cc$ efficiently} if for any
target concept $f\in C_{n,b}$,

\begin{itemize}
    \item $\algo$ runs for at most $\poly(n, \log b, 1/\epsilon, \log 1/\delta)$ steps;
    \item Any hypothesis $h$ that $\algo$ produces can be evaluated at any $x \in \bbn$ in at most
        $\poly(n, \log b, 1/\epsilon, \log 1/\delta)$ time steps.
\end{itemize}

\subsection{The functions we study}
The reader might wonder which classes of Boolean valued functions over $\bbn$
are interesting.
In this article we study classes of functions that are defined in terms
of ``$b$-literals''; these include rectangles and unions
of rectangles over $\bbn$ as well as other richer classes.
As described below, $b$-literals are a natural
extension of Boolean literals to the domain $\bbn$.

\begin{definition} A function $\ell \colon [b] \to \bits$
is a \emph{basic $b$-literal} if for some $\sigma \in \{-1,1\}$
and some $\alpha \leq \beta$  with $\alpha, \beta \in [b]$ we have
$\ell(x) = \sigma$ if $\alpha \leq x \leq \beta$, and
$\ell(x) = -\sigma$ otherwise.
A function $\ell \colon [b] \to \bits$ is a \emph{$b$-literal}
if there exists a basic $b$-literal $\ell^{\prime}$ and some fixed $z\in[b]$,
$\gcd(z,b)=1$ such that for all $x\in[b]$ we have $\ell(x)=\ell^{\prime}(xz \bmod b)$.
\end{definition}

Basic $b$-literals are the most natural extension of Boolean
literals to the domain $\bbn$.  General $b$-literals
(not necessarily basic) were previously studied in \cite{AGS} and are
also quite natural.
\begin{example} If $b$ is odd then the {\em least significant bit}
function $lsb(x)\colon [b]\to \bits$, defined by $lsb(x)=-1$
iff $x$ is even, is a $b$-literal.
\end{example}
To see this, let $z = (2)^{-1} \bmod b$ (this value exists since $b$
is odd). Let $E = \{0,2,4,\dots,b-1\}$ denote the set of all the
even residues in $[b]$, i.e. $E$ is precisely the set of inputs that
are mapped to $-1$ under $lsb.$  We have
\[E = \{0 \cdot 2, 1 \cdot
2, \dots {\frac {b-1} 2} \cdot 2\}\] and consequently
\begin{align*}
E \cdot z \bmod b &\equiv \{ 0 \cdot 2 \cdot 2^{-1} \bmod b, 1 \cdot 2 \cdot 2^{-1} \bmod b, \ldots,  \frac{b-1}{2} \cdot 2 \cdot 2^{-1}\bmod b\}\\
&\equiv \{ 0, 1, 2, \ldots, \frac{b-1}{2}\}.
\end{align*}
The function $\ell^{\prime}(x)$ which equals $-1 \mathrm{~iff~} x
\in \{0,1,\ldots \frac{b-1}{2}\}$ is a basic $b$-literal, and
consequently $lsb(x)=\ell^{\prime}(xz \bmod b)$ is a $b$-literal.

\begin{definition} A function $\fisgfunc$ is
a \emph{$k$-dimensional rectangle} if it is an \textsc{And} of $k$
basic $b$-literals $\ell_1,\dots,\ell_k$ over $k$ distinct variables
$x_{i_1},\dots,x_{i_k}$. If $f$ is a $k$-dimensional rectangle for
some $k$ then we may simply say that $f$ is a \emph{rectangle}. A
\emph{union of $s$ rectangles} $R_1, \dots, R_s$ is a function of
the form $f(x) = \textsc{Or}_{i=1}^s R_i(x).$
\end{definition}
The class of unions of $s$ rectangles over $\bbn$ is a
natural generalization of the class of $s$-term $\dnf$ over $\bnz$.
Similarly \textsc{Majority} of \textsc{Parity} of basic $b$-literals
generalizes the class of \textsc{Majority} of \textsc{Parity} of Boolean
literals, a class which has been the subject of much research
 (see e.g. \cite{JACKSON,
BRUCK, KS}).

If \textsc{G} is a logic gate with potentially unbounded fan-in
(e.g. \textsc{Majority}, \textsc{Parity}, \textsc{And}, etc.) we
write ``$s$-way \textsc{G}'' to indicate that the fan-in of
\textsc{G} is restricted to be at most $s$. Thus, for example, an
``$s$-way \textsc{Majority} of $r$-way \textsc{Parity} of
$b$-literals'' is a \textsc{Majority} of at most $s$ functions
$g_1,\dots,g_s$, each of which is a \textsc{Parity} of at most $r$
many $b$-literals. We will further assume that \emph{any two
$b$-literals which are inputs to the same gate depend on different
variables.} This is a natural restriction to impose in light of our
ultimate goal of learning unions of rectangles. Although our results
hold without this assumption, it provides simplicity in the
presentation.

\subsection{Harmonic analysis of functions over $\bbn$}\label{sec:four}
We will make use of the Fourier expansion of complex valued functions over $\bbn$.

Consider $f,g \colon \bbn \to \mathbb{C}$ endowed with the inner product $\innp{f}{g} = \E[f \conju{g}]$ and induced norm
    $\|f \|=\sqrt{\innp{f}{f}}$. Let $\pru=e^{\frac{2\pi i}{b}}$ and for each $\alpha=(\alpha_1,\ldots,\alpha_n)\in \bbn$, let
    $\chi_\alpha\colon \bbn \to \mathbb{C}$ be defined as \[\chi_\alpha(x_1,\ldots, x_n)= \pru^{\alpha_1 x_1 + \cdots + \alpha_n x_n}.\]
Let $\mathcal{B}$ denote the set of functions $\mathcal{B}=\{\chi_\alpha \colon \alpha\in\bbn \}$.
It is easy to verify the following properties:
    \begin{itemize}
        \item Elements in $\mathcal{B}$ are normal: for each $\alpha=(\alpha_1,\ldots,\alpha_n)\in \bbn$,
        we have $\|\chi_\alpha\|=1$.
        \item Elements in $\mathcal{B}$ are orthogonal: For $\alpha,\beta \in \bbn$, we have
            $\innp{\chi_\alpha}{\chi_\beta}=\left\{
            \begin{array}{c}
                1\ \mathrm{if}\ \alpha=\beta\\
                0\ \mathrm{if}\ \alpha\neq\beta
            \end{array}\right.$
        \item $\mathcal{B}$ constitutes an orthonormal basis for all functions $\{f \colon \bbn \to \mathbb{C}\}$
            considered as a vector space over $\mathbb{C}.$ Thus every $f \colon \bbn \to \mathbb{C}$ can be expressed uniquely as:
            \[ f(x)=\sum_\alpha  \fousumelt\chi_\alpha(x)
            \]
            which we refer to as the \emph{Fourier expansion} or \emph{Fourier transform} of $f$.
    \end{itemize}
The values
$\{\fousumelt \colon\alpha\in\bbn \} $ are called the \emph{Fourier coefficients}
or the \emph{Fourier spectrum} of $f$.
As is well known, \emph{Parseval's Identity} relates the values of the
coefficients to the values of the function:
\begin{lemma}[Parseval's Identity]\label{parseval}
$ \littlesum_{\alpha} |\fousumelt|^2= \E[|f|^2]$
for any $f \colon \bbn \to \mathbb{C}$.
\end{lemma}
We write $L_1(f)$ to denote $\littlesum_\alpha |\fousumelt|$ and $L_{\infty}(f)$ to denote $\max_{\alpha} |\fousumelt|$.

We will also make use of the following simple fact:
\begin{observation}\label{simplefact} For any $f, h\colon \bbn \to \mathbb{C}$ and $\dist$ over $\bbn$,
    \[|\E_{\dist} [f \conju{h}]| = | \E_{\dist}[ f \conju{\littlesum_{\alpha} \hat{h}(\alpha) \chi_\alpha}]|
        = | \littlesum_{\alpha} \conju{\hat{h}(\alpha)} \E_{\dist}[f \conju{\chi_\alpha}]|
        \leq L_1(h) \max_\alpha |\E_{\dist}[f \conju{\chi_\alpha}]|.\]
\end{observation}

\subsection{Additional tools:  weak hypotheses and boosting}
\begin{definition} Let $\fisgfunc$ and $\dist$ be a probability
distribution over $\bbn$. A function $g \colon \bbn \to [-1,1]$ is
said to be a \emph{weak hypothesis for $f$ with advantage $\gamma$
under $\dist$} if $\E_{\dist}[f g] \geq \gamma$.
\end{definition}

The first {\em boosting} algorithm was described by Schapire
\cite{SCH} in 1990; since then boosting has been intensively studied
(see \cite{FS99} for an overview).  The basic idea is that by
combining a sequence of weak hypotheses $h_1,h_2,\dots$ (the $i$-th
of which has advantage $\gamma$ with respect to a carefully chosen
distribution $\dist_i$) it is possible to obtain a high accuracy
final hypothesis $h$ which satisfies $\Pr[h(x) = f(x)] \geq 1-\eps.$
The following theorem, which can be obtained easily from the results
of \cite[Section 2.3]{Servedio:03jmlr}, gives a precise statement
of the performance guarantees of a particular boosting algorithm,
which we call Algorithm $\mathcal{B}$.  Many similar statements are
now known about a range of different boosting algorithms but this is
sufficient for our purposes.
\begin{theorem}[Boosting Algorithm \cite{Servedio:03jmlr}]\label{boosting}
    Suppose that Algorithm $\mathcal{B}$ is given:
    \begin{itemize}
        \item $0<\epsilon,\delta<1$, and membership query access $\mq$ to
$\fisgfunc$;
        \item access to
an algorithm \texttt{WL} which has the following property:
given a value $\delta'$ and access to $\mq$ and to $\exo$ (the latter is an
example oracle which generates random examples from $\bbn$ drawn
with respect to distribution $\dist$), it constructs a weak
hypothesis for $f$ with advantage $\gamma$ under $\dist$ with probability at
least $1-\delta^{\prime}$ in time polynomial in $n$, $\log b$,
$\log (1/\delta^{\prime})$.
\end{itemize}
Then Algorithm $\mathcal{B}$ behaves as follows:
\begin{itemize}
\item It runs for $S=O(1/\epsilon\gamma^2)$ stages and runs
in total time polynomial in $n$, $\log b$, ${\epsilon}^{-1}$,
${\gamma}^{-1}$, $\log ({\delta}^{-1})$.

\item At each stage $1\leq j \leq S$ it constructs a distribution $\dist_j$
such that $L_{\infty}(\dist_{j})< \poly({\epsilon}^{-1})/b^n$, and
simulates $\mathsf{EX}(f,\mathscr{D}_j)$ for \texttt{WL} in stage
$j$. Moreover, there is a value $c \in [1/2, 3/2]$ (the precise
value of $c$ depends on $\mathscr{D}_j$ and is not known to the
algorithm) and a fixed ``pseudo-distribution''
$\tilde{\mathscr{D}}_j$ satisfying $\tilde{\mathscr{D}}_j(x) = c
\mathscr{D}_j(x)$ for all $x$, such that $\tilde{\dist}_{j}(x)$ can
be computed in time polynomial in $n \log b$ for each $x\in \bbn$.

\item It outputs a final hypothesis $h=\sign(h_1+h_2+\ldots+h_{S})$
which $\epsilon$-approximates $f$ under the
uniform distribution with probability $1-\delta$; here $h_j$ is the output
of \texttt{WL} at stage $j$ invoked with simulated access to
$\mathsf{EX}(f,\mathscr{D}_{j})$.
\end{itemize}
\end{theorem}

We will sometimes informally refer to distributions $\dist$ which satisfy the bound
$L_{\infty}(\dist)< \frac{\poly({\epsilon}^{-1})}{b^n}$ as \emph{smooth} distributions.

In order to use boosting, it must be the case that there exists a
suitable weak hypothesis with advantage $\gamma$. In this paper we
will use the ``discriminator lemma'' of Hajnal \emph{et al.}
\cite{HMPST} (see also \cite{Pis81}) at various points (see e.g. the
proofs of Theorem~\ref{mthm1} and Lemma~\ref{lemma1app2}) to assert
that the desired weak hypothesis exists:

\begin{lemma}[The Discriminator Lemma \cite{HMPST,Pis81}]\label{disclemma}Let $\mathfrak{H}$ be a class of $\pm 1$-valued functions over
$\bbn$ and let $\fisgfunc$ be expressible as $$f=\textsc{Majority}
(h_1,\ldots,h_s)$$ where each $h_i\in\mathfrak{H}$ and
    $h_1(x)+\ldots+h_s(x)\neq 0$ for all x. Then for any distribution $\dist$ over $\bbn$ there is some $h_i$ such that
    $|\E_{\dist}[f h_i]|\geq 1/s$.
\end{lemma}

\section{The Generalized Harmonic Sieve Algorithm} \label{sec:GHS}

In this section our goal is to describe a variant of Jackson's Harmonic Sieve
Algorithm and show that under suitable conditions it can efficiently
learn certain functions $\fisgfunc$.
As mentioned earlier, our aim is to attain $\poly(\log b)$
runtime dependence on $b$ and consequently obtain efficient algorithms
as described in Section~\ref{sec:prelim}.
This goal precludes using Jackson's original Harmonic Sieve variant for
$\bbn$ since the runtime of his weak learner depends polynomially
rather than polylogarithmically on $b$ (see
\cite[Lemma 15]{JACKSON}).

As we describe below, this $\poly(\log b)$ runtime
can be achieved by modifying the Harmonic Sieve over $\bbn$ to use
a weak learner due
to Akavia {\em et al.}~\cite{AGS} which is more efficient
than Jackson's weak learner.
We shall call the resulting algorithm ``The Generalized
Harmonic Sieve'' algorithm, or $\ghs$ for short.

Recall that in the Harmonic Sieve over the Boolean domain $\bits^n$,
the weak hypotheses used are simply the Fourier basis elements over
$\bits^n$, which correspond to the Boolean-valued parity functions.
For $\bbn$, we will use the real component of the complex-valued
Fourier basis elements $\{\chi_\alpha, \alpha \in \bbn\}$ (as defined in
Section~\ref{sec:four}) as our weak hypotheses.

The following theorem of Akavia \emph{et al.}~\cite[Theorem 5]{AGS} will play a crucial role
towards construction of the $\ghs$ algorithm.
\begin{theorem}[See \cite{AGS}]\label{sec3thm1}\hspace*{-1pt}There is a learning algorithm that, given membership query access to $f \colon \bbn \to \mathbb{C}$, $0<\gamma$ and $0<\delta<1$,
outputs a list $L$ of indices such that with probability at least $1 - \delta$, we have
$\{\alpha \colon |\hat{f}(\alpha)|>\gamma\} \subseteq L$ and $|\hat{f}(\beta)| \geq {\frac \gamma 2}$ for every $\beta \in L.$  The
    running time of the algorithm is polynomial in $n$, $\log b$, $\|f\|_{\infty}$, $\gamma^{-1}$, $\log (\delta^{-1})$.
\end{theorem}

\begin{lemma}[Construction of the weak hypothesis]\label{sec3lemma1} Given
    \begin{itemize}
        \item Membership query access $\mq$ to $\fisgfunc$;
        \item A smooth distribution $\dist$; more precisely,
access to an algorithm computing $\tilde{\dist}(x)$ in
time polynomial in $n$, $\log b$ for each $x\in \bbn$.
Here $\tilde{\dist}$ is a ``pseudo-distribution'' for $\dist$
as in Theorem~\ref{boosting}, i.e. there is a value $c \in [1/2,3/2]$
such that $\tilde{\dist}(x)=c\dist(x)$ for all $x.$
        \item A value
        $0< \gamma<1/2$ such that there exists an element of the Fourier basis $\chi_\tau$ satisfying
            $|\E_{\dist}[f \conju{\chi_\tau}]| > \gamma$,
    \end{itemize}
    there is an algorithm that outputs a weak hypothesis for $f$ with advantage $\gamma/4$ under $\dist$
    with probability $1-\delta$ and runs in time polynomial in $n$, $\log b$, $\epsilon^{-1}$,
    $\gamma^{-1}$, $\log (\delta^{-1})$.
\end{lemma}
\begin{pf} Let $f_{\ast}(x)=b^n \tilde{\dist}(x) f(x)$. Observe that
    \begin{itemize}
        \item Since $\dist$ is smooth, $\|f_{\ast}\|_{\infty}<\poly({\epsilon}^{-1})$.
        \item For any $\alpha\in\bbn$,
    $\hat{f_{\ast}}(\alpha)= \E[f_{\ast} \conju{\chi_\alpha} ] = \frac{1}{b^n} \displaystyle\littlesum_{x\in \bbn} b^n \tilde{\dist}(x) f(x) \conju{\chi_\alpha(x)}=
 \E_{\dist}[cf \conju{\chi_\alpha}]$.
    \end{itemize}
    Therefore one can invoke the algorithm of Theorem~\ref{sec3thm1} over $f_{\ast}(x)$ by simulating
    $\mathsf{MEM}(f_{\ast})$ via $\mq$, each time
    with $\poly(n,\log b)$ time overhead, and obtain a list $L$ of indices.
    Note that since we are guaranteed that there exists an index $\tau$ satisfying $|\E_{\dist}[f \conju{\chi_\tau}]| > \gamma$
    implying $|\hat{f_{\ast}}(\tau)|\geq c \gamma$, we can invoke Theorem~\ref{sec3thm1} in such a way that for any index $\beta$
    in its output, we know $|\hat{f_{\ast}}(\beta)|\geq c \gamma/2$.

    It is easy to see that the algorithm runs in the desired
    time bound and outputs a nonempty list $L$. Let $\beta$ be any element of $L$.
    Since $\hat{f_{\ast}}(\beta)=\E[b^n \tilde{\dist}(x) f(x)
    \conju{\chi_\beta(x)}]$,
    one can approximate $\frac{\E_{\dist}[f \conju{\chi_\beta}]}{|\E_{\dist}
    [f \conju{\chi_\beta}]|}=
    \frac{\hat{f_{\ast}}(\beta)}{|\hat{f_{\ast}}(\beta)|}=e^{i\theta}$ using uniformly drawn random examples.
    Let $e^{i{\theta}^\prime}$ be the approximation thus obtained.

    By assumption we know that for random $x \in [b]^n$, the random variable $$(b^n
        \tilde{\dist}(x) f(x) \conju{\chi_\beta(x)})$$ always takes
        a value whose magnitude is $O(\poly({\epsilon}^{-1}))$
            in absolute value.
    Using a straightforward Chernoff bound argument, this implies that $|\theta-{\theta}^{\prime}|$ can be made
    smaller than any constant using $\poly(n, \log b,\epsilon^{-1})$ time
    and random examples.

Now observe that we have
\[
\E_{\dist} [f \conju{\chi_{\beta}}]= e^{i\theta} |\E_{\dist} [f \conju{\chi_{\beta}}]| \Rightarrow
        \E_{\dist} [f \conju{e^{i \theta} \chi_{\beta}}]=|\E_{\dist} [f \conju{\chi_{\beta}}]| = c^{-1}|\hat{f_{\ast}}(\beta)|\geq \gamma/2.
\]

Therefore for a sufficiently small value of $|\theta-{\theta}^{\prime}|$, we have
\[
\E_{\dist} [f \Re \{\conju{e^{i {\theta}^{\prime}} \chi_{\beta}}\}]=\Re \{ \E_{\dist} [f \conju{e^{i {\theta}^{\prime}} \chi_{\beta}}]\}
=\Re \{ e^{i (\theta-{\theta}^{\prime})}\underbrace{\E_{\dist} [f \conju{e^{i {\theta}} \chi_{\beta}}]}_{\text{real valued and $\geq\gamma/2$}}\} \geq \gamma/4.
\]
Since $\Re \{\conju{e^{i {\theta}^{\prime}} \chi_{\beta}}\}$ always
takes values in $[-1,1]$, we conclude that $\Re \{\conju{e^{i
{\theta}^{\prime}} \chi_{\beta}}\}$ constitutes a weak hypothesis
for $f$ with advantage $\gamma/4$ under $\dist$ with high
probability. \qed
\end{pf}
Rephrasing the statement of Lemma~\ref{sec3lemma1}, now we know:
As long as for any function $f$ in the concept class
it is guaranteed that under any smooth distribution
$\dist$ there is a Fourier basis element $\chi_\beta$
that has non-negligible correlation with $f$  (i.e. $|\E_{\dist}
[f \conju{\chi_\alpha}]| > \gamma$), then it is possible
to efficiently identify and use such a Fourier basis element
to construct a weak hypothesis.

Now one can invoke Algorithm $\mathcal{B}$ from Theorem~\ref{boosting} as in Jackson's original Harmonic Sieve: At stage $j$, we have a distribution $\dist_j$ over $\bbn$
for which $L_{\infty}(\dist_{j})< \poly(\epsilon^{-1})/b^n$. Thus one can pass the values of $\dist_j$ to the algorithm in
Lemma~\ref{sec3lemma1} and use this algorithm as \texttt{WL} in Algorithm $\mathcal{B}$ to obtain the weak hypothesis at each stage.
Repeating this idea for every stage and combining the weak hypotheses generated for all the stages as described by Theorem~\ref{boosting}, we have the $\ghs$ algorithm:
\begin{corollary}[The Generalized Harmonic Sieve]\label{ghscor} Let $\mathfrak{C}$
be a concept class. Suppose that for any concept $f \in C_{n,b}$ and
    any distribution $\dist$ over $\bbn$ with $L_{\infty}(\dist)<
    \poly({\epsilon}^{-1})/b^n$ there exists a Fourier basis element
$\chi_\alpha$ such that $|\E_{\dist}[f \conju{\chi_\alpha}]|$
$\geq \gamma$.
Then $\mathfrak{C}$ can be learned in time
$\poly(n, \log b, {\epsilon}^{-1}, {\gamma}^{-1})$.
\end{corollary}

\section{Learning {\sc Majority} of {\sc Parity} using $\ghs$}
\label{sec:firstapproach}

In this section we identify classes of functions which can be
learned efficiently using the $\ghs$ algorithm and prove Theorem~\ref{thmapp1}.

Let $\cc^{\circ}$ denote the concept class of Theorem~\ref{thmapp1}:
the concept class of $s$-way \textsc{Majority} of $r$-way
\textsc{Parity} of $b$-literals where $s=\poly (n \log b)$,
$r=O(\frac{\log(n \log b)} {\log \log (n \log b)})$.

To prove Theorem~\ref{thmapp1}, we show that for any concept $f \in \cc^{\circ}$ and
under any smooth distribution there must be some Fourier basis element which
has high correlation with $f$; this is the essential step which lets us apply
the Generalized Harmonic Sieve. We prove this in Section~\ref{sec:weakbasis}.
In Section~\ref{sec:better} we give an alternate argument which yields
a Theorem~\ref{thmapp1} analogue but with a slightly different bound on $r$, namely
$r=O(\frac{\log(n \log b)} {\log \log b})$.

\subsection{Setting the stage} \label{sec:setstage}

In this section we first focus our attention to functions defined over $[b]$, i.e. the case $n=1$.

For ease of notation we will write
$abs(\alpha)$ to denote $\min\{\alpha, b-\alpha \}$.  We will use the following simple lemma from
\cite{AGS}:
\begin{lemma}[See \cite{AGS}]\label{recsum} For all $0\leq\ell\leq b$, we have
$|\littlesum^{\ell-1}_{y=0} \pru^{\alpha y}| < b/abs(\alpha)$.
\end{lemma}

\begin{corollary}\label{litbnd} Let $f\colon [b] \to \bits$ be a basic $b$-literal.  Then if $\alpha=0$, $|\fousumelt|\leq1$, while
if $\alpha\neq 0$,    $|\fousumelt| < \frac{2}{abs(\alpha)}$.
\end{corollary}
\begin{pf}
The first inequality follows immediately from Parseval's Identity given in Lemma~\ref{parseval},
because $f$ is $\{1, -1\}$-valued. For the latter, note that
$|\hat{f}(\alpha)|=|\E[f \conju{\chi_{\alpha}}]|=$
\[\frac{1}{b}
\left|
\littlesum_{x\in f^{-1}(1)} \chi_{\alpha}(x) - \littlesum_{x
\in f^{-1}(-1)}  \chi_{\alpha}(x) \right|
\leq {\frac 1 b}
\left|\littlesum_{x\in f^{-1}(1)} \chi_{\alpha}(x)\right|
+ {\frac 1 b}
\left|\littlesum_{x
\in f^{-1}(-1)}  \chi_{\alpha}(x) \right|
\]
where the inequality is simply the triangle inequality.
It is easy to see that each of the sums on the RHS above equals
$
{\frac 1 b} \left|\pru^{\alpha c}\right|
|\littlesum^{\ell-1}_{y=0} \pru^{\alpha y}|
=
{\frac 1 b} |\littlesum^{\ell-1}_{y=0} \pru^{\alpha y}|
$
for some suitable $c$ and $\ell \leq b$, and hence Lemma~\ref{recsum}
gives the desired result.
\qed
\end{pf}

The following easy lemma is useful for relating the Fourier transform of a
$b$-literal to the corresponding basic $b$-literal:
\begin{lemma} \label{lem:bbasic}
For $f,g \colon [b] \to \mathbb{C}$ such that $g(x)=f(xz)$ where $\gcd(z,b)=1$, we have $\hat{g}(\alpha)=\hat{f}(\alpha z^{-1})$.
\end{lemma}
\begin{pf}
    \begin{align*}
    \hat{g}(\alpha)& = \E_{x}[g(x)\conju{\chi_\alpha(x)}]=\E_{x}[f(xz)\conju{\chi_\alpha(x)}]=\E_{xz^{-1}}[f(x)\conju{\chi_\alpha(xz^{-1})}]\\
    &=\E_{xz^{-1}}[f(x)\conju{\chi_{\alpha z^{-1}}(x)}]=\E_{x}[f(x)\conju{\chi_{\alpha z^{-1}}(x)}]= \hat{f}(\alpha z^{-1}).  \quad
     \qed
\end{align*}
\end{pf}

A natural way to approximate a $b$-literal is by truncating its Fourier representation.  We make
the following definition:

\begin{definition} Let $k$ be a positive integer. For $f\colon [b] \to \bits$ a basic $b$-literal, the \emph{$k$-restriction of $f$} is $\tilde{f}\colon [b] \to \mathbb{C},$
$\tilde{f}(x)=\littlesum_{abs(\alpha)\leq k} \hat{f}(\alpha) \chi_{\alpha}(x).$
More generally, for $f\colon [b] \to \bits$ a $b$-literal (so
$f(x) = f'(xz)$ where $f'$ is a basic $b$-literal) the
\emph{$k$-restriction of $f$} is $\tilde{f}\colon [b] \to \mathbb{C},$
$\tilde{f}(x)=\littlesum_{abs(\alpha z^{-1})\leq k} \hat{f}(\alpha) \chi_{\alpha}(x) = \littlesum_{abs(\beta) \leq k} \widehat{f'}(\beta) \chi_\beta(x z).$
\end{definition}

\subsection{There exist highly correlated Fourier basis elements for functions in
$\cc^{\circ}$ under smooth distributions} \label{sec:weakbasis}

In this section we show that given any $f \in \cc^{\circ}$, the
concept class of Theorem~\ref{thmapp1}, and any smooth distribution
$\dist$, some Fourier basis element must have high correlation with
$f$.  In more detail, the main result of this section is the
following theorem:

\begin{theorem}\label{mthm1} Let $\tau \geq 1$ be any value, and let
$\cc$ be the concept class consisting of
    $s$-way \textsc{Majority} of $r$-way \textsc{Parity} of $b$-literals where $s=\poly (\tau)$ and
    $r=O(\frac{\log(\tau)} {\log \log(\tau)})$. Then for any $f\in C_{n,b}$ and any distribution $\dist$ over
    $\bbn$ with $L_{\infty}(\dist)=\poly (\tau)/b^n$, there exists a Fourier basis element $\chi_\alpha$ such that
$$|\E_{\dist}[f \conju{\chi_\alpha}]|>\Omega(1/\poly(\tau)).$$
\end{theorem}

We prove the theorem after some preliminary lemmata about
approximating basic $b$-literals and products of basic $b$-literals.
We begin by bounding the error of the $k$-restriction of a basic
$b$-literal:

\begin{lemma}\label{lem1} For $f\colon [b] \to \bits$ a $b$-literal and $\tilde{f}$ the $k$-restriction
of $f$, we have $\E[|f-\tilde{f}|^2] = 8/k$ and $\E[|f-\tilde{f}|]<\sqrt{8/k}$.
\end{lemma}
\begin{pf} Without loss of generality assume $f$ to be a basic $b$-literal. By an immediate application of Lemma~\ref{parseval}
    (Parseval's Identity) we obtain:
    \[\E[|f-\tilde{f}|^2] = \littlesum_{abs(\alpha)> k} |\hat{f}(\alpha)|^2 \ \ \underbrace{<}_{\mathrm{by}\ \mathrm{Corollary}\ \ref{litbnd}} \ \ 2 \cdot \littlesum_{m= k+1}^{\infty} \frac{4}{m^2} < 8 \int_{k}^{\infty} \frac{1}{\xi^2} d\xi = \frac{8}{k}.
     \]
     By the non-negativity of variance, this implies $\E[|f-\tilde{f}|]<\sqrt{8/k}. \qed$
\end{pf}

Now suppose that $f$ is an $r$-way \textsc{Parity} of $b$-literals
$f_1,\dots,f_r$.  Since \textsc{Parity} corresponds to
multiplication over the domain $\{-1,1\}$, this means that $f=
\prod^{r}_{i=1} f_i$.  It is natural to approximate $f$ by the
product of the $k$-restrictions $\prod^{r}_{i=1} \tilde{f}_i$. The
following lemma bounds the error of this approximation:
\begin{lemma}\label{mlemma} For $i=1,\ldots, r$, let $f_i \colon [b] \to \bits$ be a $b$-literal and let $\tilde{f}_i$ be its $k$-restriction. Then
    $$\E[|f_1(x_1) f_2(x_2)\ldots f_r(x_r) - \tilde{f}_1(x_1) \tilde{f}_2(x_2)\ldots \tilde{f}_r(x_r)| ]< e^{r \sqrt{8/k}}-1.$$
\end{lemma}

\begin{pf} First note that by Lemma~\ref{lem1},
we have that for each $i=1,\ldots, r$:
    \[\E_{x_i}[|f_i(x_i) - \tilde{f}_i(x_i)|]
    \leq \sqrt{\E_{x_i}[|f_i(x_i) - \tilde{f}_i(x_i)|^2]} < \sqrt{8/k}.\]
    Therefore we also have for each $i=1,\ldots, r$:
    \[\E_{x_i}[|\tilde{f}_i(x_i)|]< \underbrace{\E_{x_i}[|\tilde{f}_i(x_i) - f_i(x_i)|]}_{<\sqrt{8/k}} +
    \underbrace{\E_{x_i}[|f_i(x_i)|]}_{=1} < 1+ \sqrt{8/k}.\]
For any $(x_1,\ldots,x_r)$ we can bound the difference in the lemma as follows:
    \begin{eqnarray*}
&&|f_1(x_1)\ldots f_r(x_r) - \tilde{f}_1(x_1)\ldots \tilde{f}_r(x_r) |\leq \\
&&|f_1(x_1)\ldots f_r(x_r) - f_1(x_1)\ldots f_{r-1}(x_{r-1})\tilde{f}_r(x_r)|
+\\
&&|f_1(x_1)\ldots f_{r-1}(x_{r-1})\tilde{f}_r(x_r) - \tilde{f}_1(x_1)\ldots \tilde{f}_r(x_r) |\leq\\
&&
|f_r(x_r) - \tilde{f}_r(x_r)| +
        |\tilde{f}_r(x_r)| |f_1(x_1) \ldots f_{r-1}(x_{r-1}) - \tilde{f}_1(x_1) \ldots \tilde{f}_{r-1}(x_{r-1}) |
\end{eqnarray*}
    Therefore the expectation in question is at most:
    \scriptsize
    \begin{align*}
&\underbrace{\underset{x_r}{\E}[|f_r(x_r) - \tilde{f}_r(x_r)|]}_{<\sqrt{8/k}} + \underbrace{\underset{x_r}{\E}[|\tilde{f}_r(x_r)|]}_{< 1+\sqrt{8/k}} \cdot \E_{(x_1,\ldots,x_{r-1})}[|f_1(x_1)\ldots f_{r-1}(x_{r-1}) - \tilde{f}_1(x_1) \ldots \tilde{f}_{r-1}(x_{r-1})|].
\end{align*}
\normalsize
We can repeat this argument successively until the base case
$$\E_{x_1}[|f_1(x_1) - \tilde{f}_1(x_1)|]< {\sqrt{8/k}}$$ is reached.
Thus one obtains the upper bound
\begin{align*}
\E[|f_1(x_1) \ldots f_r(x_r) - \tilde{f}_1(x_1) \ldots \tilde{f}_r(x_r)|]&<
\sqrt{8/k} \sum_{i=0}^{r-1} (1+\sqrt{8/k})^{i}\\
&=(1+\sqrt{8/k})^{r}-1 < e^{r \sqrt{8/k}}-1.
\qed
\end{align*}
\end{pf}

Now we are ready to prove Theorem~\ref{mthm1}, which asserts the
existence (under suitable conditions) of a highly correlated Fourier
basis element. The basic approach of the following proof is
reminiscent of the main technical lemma from \cite{JKS}.

\begin{pf*}{PROOF OF THEOREM~\ref{mthm1}.}
    Assume $f$ is a \textsc{Majority} of $h_1,\ldots,h_s$ each of which is a $r$-way
    \textsc{Parity} of $b$-literals.
    Then Lemma~\ref{disclemma} implies that there exists $h_i$ such that $|\E_{\dist}[f h_i]| \geq 1/s$. Let $h_i$
    be \textsc{Parity} of the $b$-literals $\ell_1, \ldots, \ell_r$.

    Since $s$ and $b^n \cdot L_{\infty}(\dist)$ are
    both at most $\poly (\tau)$ and $r=O(\frac{\log(\tau)} {\log \log(\tau)})$, Lemma~\ref{mlemma} implies that
    there are absolute constants $C_1,C_2$ such that if we consider the $k$-restrictions
    $\tilde{\ell}_1,\dots,\tilde{\ell}_r$ of $\ell_1,\ldots, \ell_r$ for $k=C_1 \cdot \tau^{C_2}$, we will have
$\E[|h_i - \littleprod_{j=1}^r\tilde{\ell}_j|]\leq 1/(2 s b^n L_{\infty}
(\dist))$
where the expectation on the left hand side is with respect to the
uniform distribution on $\bbn$.  This in turn implies that
$\E_{\dist}[|h_i - \littleprod_{j=1}^r \tilde{\ell}_j|]\leq 1/2s.$
Let us write $h^{\prime}$ to denote $\littleprod_{j=1}^r \tilde{\ell}_j$.
We then have
    \begin{align*}
        |\E_{\dist} [f \conju{h^{\prime}}]| & \geq |\E_{\dist} [f \conju{h_i}]| - | \E_{\dist} [f \conju{(h_i - h^{\prime})}]|
 \geq  |\E_{\dist} [f \conju{h_i}]| - \E_{\dist}[|f\conju{(h_i - h^{\prime})}|]\\
 &=  |\E_{\dist} [f h_i]| - \E_{\dist}[|h_i - h^{\prime}|]
        \geq 1/s - 1/2s = 1/2s.
    \end{align*}
    By Observation~\ref{simplefact} we additionally have
    $$
        |\E_{\dist} [f \conju{h^{\prime}}]| = | \leq L_1(h^{\prime}) \max_\alpha |\E_{\dist}[f \conju{\chi_\alpha}]|.
    $$
Moreover, for each $j=1,\ldots,r$ we have the following
(where we write $\ell'_j$ to denote the
basic $b$-literal associated with the $b$-literal $\ell_j$):
    \[
    L_1(\tilde{\ell}_j) = \littlesum_{abs(\alpha)\leq k} |\widehat{\ell'}_j(\alpha)| \underbrace{<}_{
    \mathrm{by}\ \mathrm{Corollary}\ \ref{litbnd}} 1 + 2 \littlesum_{m=1}^{k} 2/m < 5 + 4 \ln (k+1).
    \]
Therefore, for some absolute constant $c>0$ we have
$L_1(h^{\prime})\leq \prod^{r}_{j=1} L_1(\tilde{\ell}_j) \leq (c
\log k)^r$, where the first inequality holds as a consequence of the
elementary fact that the $L_1$ norm of a product is at most the
product of the $L_1$ norms of the components. Combining inequalities, we obtain
    \[\max_\alpha |\E_{\dist}[f \conju{\chi_\alpha}]|\geq 1/(2s(c \log k)^r) = \Omega(1/\poly(\tau))\]
which is the desired result. \qed
\end{pf*}

Since we are interested in algorithms with runtime $\poly(n, \log b,
{\epsilon}^{-1})$, setting $\tau=n{\epsilon}^{-1}\log b$ in
Theorem~\ref{mthm1} and combining its result with
Corollary~\ref{ghscor}, gives rise to Theorem~\ref{thmapp1}.

\subsection{The second approach}
\label{sec:better}

A different analysis, similar to that which Jackson uses in the proof of
\cite[Fact 14]{JACKSON}, gives us an alternate bound to Theorem~\ref{mthm1}:

\begin{lemma}\label{lemma1app2} Let $\cc$ be the concept class consisting of $s$-way
\textsc{Majority} of $r$-way \textsc{Parity} of $b$-literals.
    Then for any $f\in C_{n,b}$ and any distribution $\dist$ over $\bbn$, there exists a Fourier basis element $\chi_\alpha$
    such that
$|\E_{\dist}[f \conju{\chi_\alpha}]|=\Omega(1/s{(\log b)}^r).$
\end{lemma}
\begin{pf}
    Assume $f$ is a \textsc{Majority} of $h_1,\ldots,h_s$ each of which is a $r$-way
    \textsc{Parity} of $b$-literals. Then
    Lemma~\ref{disclemma} implies that there exists $h_i$ such that $|\E_{\dist}[f h_i]| \geq 1/s$. Let $h_i$ be
    \textsc{Parity}
    of the $b$-literals $\ell_1, \ldots, \ell_r$.
    Observation~\ref{simplefact} gives:
\[
        1/s\leq|\E_{\dist}[f h_i]|=|\E_{\dist} [f \conju{h_i}]|
        \leq L_1(h_i) \max_\alpha |\E_{\dist}[f \conju{\chi_\alpha}]|
\]
Also note that for $j=1,\ldots,r$ we have the following (where as before
we write $\ell'_j$ to denote the basic $b$-literal associated with the $b$-literal $\ell_j$):
    \[
    L_1(\ell_j) \  \underbrace{=}_{\mathrm{by}\ \mathrm{Lemma}\ \ref{lem:bbasic}} \ \littlesum_{\alpha} |\widehat{\ell'_j}(\alpha)| \ \underbrace{<}_{\mathrm{by}\ \mathrm{Corollary}\ \ref{litbnd}} \ 1 +  2 \cdot \littlesum_{m=1}^{b-1} 2/m < 5 + 4\ln b.
    \]
    Therefore for some constant $c>0$ we have $L_1(h_i)\leq \prod^{r}_{j=1} L_1(\ell_j) =O((\log b)^r)$, from which
    we obtain $\max_\alpha |\E_{\dist}[f \conju{\chi_\alpha}]|=\Omega(1/s{(\log b)}^r)$.
\qed
\end{pf}
Combining this result with that of Corollary~\ref{ghscor} we obtain the following result:
\begin{theorem}\label{mthm2} The concept class $\cc$ consisting of $s$-way
\textsc{Majority} of $r$-way \textsc{Parity} of $b$-literals can be
learned in time $\poly(s, n ,(\log b)^r)$ using the $\ghs$
algorithm.
\end{theorem}
As an immediate corollary we obtain the following close analogue of Theorem~\ref{thmapp1}:
\begin{theorem}\label{thm:analogue}
The concept class $\cc$ consisting of $s$-way \textsc{Majority} of
$r$-way \textsc{Parity} of $b$-literals where $s=\poly (n \log b)$,
$r=O(\frac{\log(n \log b)} {\log \log b})$ is efficiently learnable
using the $\ghs$ algorithm.
\end{theorem}

\section{Locating sensitive elements and learning with $\ghs$ on a restricted grid}
 \label{sec:grid}

In this section we consider an extension of the $\ghs$ algorithm which lets
us achieve slightly better bounds when we are dealing only with basic $b$-literals.
Following an idea from \cite{BK}, the new algorithm works by identifying a subset of
``sensitive'' elements from $[b]$ for each of the $n$ dimensions. 

\begin{definition}[See \cite{BK}] A value $\sigma\in[b]$ is called \emph{$i$-sensitive} with respect to $f\colon \bbn$ $\to \bits$
    if there exist values $c_1,c_2,\ldots,c_{i-1},c_{i+1},\ldots,c_n\in [b]$ such that
    $$f(c_1,\ldots,c_{i-1},(\sigma-1) \bmod b,c_{i+1},\ldots,c_n)\neq f(c_1,\ldots,c_{i-1},\sigma,c_{i+1},\ldots,c_n).$$
    A value $\sigma$ is called \emph{sensitive} with respect to $f$ if $\sigma$ is $i$-sensitive for some $i$.
    If there is no $i$-sensitive value with respect to $f$, we say index $i$ is \emph{trivial}.
\end{definition}
The main idea is to run $\ghs$ over a restricted subset of the original domain $\bbn$, which is the grid
formed by the sensitive values and a few more additional values, and therefore lower the algorithm's complexity.
\begin{definition} A \emph{grid in $\bbn$} is a set $\mathscr{S}=L_1 \times L_2 \times \cdots \times L_n$
    with $0 \in  L_i \subseteq [b]$ for each $i$. We refer to the elements of $\mathscr{S}$ as {\em corners}.
    The \emph{region covered by a corner} $(x_1,\ldots,x_n)\in \mathscr{S}$ is defined to be the set
    $\{(y_1,\ldots,y_n)\in\bbn \colon \forall i, x_i\leq y_i<\lceil x_i\rceil\}$ where $\lceil x_i\rceil$
    denotes the smallest value in $L_i$ which is larger than $x_i$ (by convention
    $\lceil x_i\rceil:=b$ if no such value exists). The \emph{area covered by the corner} $(x_1,\ldots,x_n)\in \mathscr{S}$
    is therefore defined to be $\littleprod_{i=1}^n (\lceil x_i \rceil - x_i)$.
    A \emph{refinement} of $\mathscr{S}$ is a grid in $\bbn$ of the form
    $L^{\prime}_{1} \times L^{\prime}_{2} \times \cdots \times L^{\prime}_{n}$ where each $L_i\subseteq L^{\prime}_{i}$.
\end{definition}

\begin{lemma}\label{lemma2app2} Let $\mathscr{S}$ be a grid $L_1 \times L_2 \times \cdots \times L_n$ in $\bbn$ such that
    each $|L_i|\leq \ell$. Let $\mathcal{I}_\mathscr{S}$ denote the set of indices for which $L_i\neq\{0\}.$
    If $|\mathcal{I}_\mathscr{S}|\leq \kappa$, then $\mathscr{S}$ admits a refinement
    $\mathscr{S}^{\prime}=L^{\prime}_{1} \times L^{\prime}_{2} \times \cdots \times L^{\prime}_{n}$ such that
    \begin{enumerate}
        \item All of the sets $L'_i$ which contain
        more than one element have the same number of elements: $\mathbf{L_{max}}$, which is
        at most $\ell + C \kappa \ell$, where $C={\frac b {\kappa\ell}} \cdot {\frac 1 {\lfloor b/
        4 \kappa \ell \rfloor}}  \geq 4.$
        \item Given a list of the sets $L_1,\dots,L_n$ as input, a list of the sets
        $L'_1,\dots,L'_n$ can be generated by an algorithm with a running time of $O(n \kappa \ell \log b)$.
        \item $L^{\prime}_{i}=\{0\}$ whenever $L_i=\{0\}$.
        \item Any $\epsilon$ fraction of the corners in
            $\mathscr{S}^{\prime}$ cover a combined area of at most $2 \epsilon b^n$.
    \end{enumerate}
\end{lemma}
\begin{pf} Consider Algorithm $1$ which, given $\mathscr{S}=L_1 \times L_2 \times \cdots \times L_n$,
    generates $\mathscr{S}^{\prime}$.

    The purpose of the code between lines 18--22 is to make every $L^{\prime}_i\neq \{0\}$ contain equal
    number of elements. Therefore the algorithm keeps track of the number of elements in the largest
    $L^{\prime}_i$ in a variable called $\mathbf{L_{max}}$ and eventually adds more (arbitrary)
     elements to those
    $L^{\prime}_i\neq \{0\}$ which have fewer elements.
\begin{algorithm}[t]
  \caption{Computing a refinement of the grid $\mathscr{S}$ with the desired properties.}
  \begin{algorithmic}[1]
      \STATE $\mathbf{L_{max}}\leftarrow 0$.
      \FORALL{$1\leq i \leq n$}
      \IF{$L_i=\{0\}$}
      \STATE $L^{\prime}_i \leftarrow\{0\}$.
      \ELSE
      \STATE Consider $L_i = \{ x^{i}_{0},  x^{i}_{1}, \ldots, x^{i}_{\ell-1}\}$, where
      $x^{i}_{0}< x^{i}_{1}< \cdots  < x^{i}_{\ell-1}$ (Also let $x^{i}_{\ell}=b$).
      \STATE Set $L^{\prime}_i\leftarrow L_i$ and
      $\tau\leftarrow\lfloor b/4\kappa\ell \rfloor$.
      \FORALL{$r=0,\dots,\ell-1$}
      \IF{$|x^{i}_{r+1} - x^{i}_{r}| >  \tau $}
      \STATE $L^{\prime}_i \leftarrow L^{\prime}_i \cup \{ x^{i}_{r}+ \tau,  x^{i}_{r}+ 2\tau, \ldots \}$
      (up to and including the largest $x^{i}_{r}+ j\cdot\tau$ which is less than $x^i_{r+1}$)
      \ENDIF
      \ENDFOR
      \IF{$|L^{\prime}_{i}|> \mathbf{L_{max}}$}
      \STATE $\mathbf{L_{max}}\leftarrow |L^{\prime}_{i}|$.
      \ENDIF
      \ENDIF
      \ENDFOR
      \FORALL{$1\leq i \leq n$ with $|L^{\prime}_i|>1$}
      \WHILE{$(|L^{\prime}_i|< \mathbf{L_{max}})$}
      \STATE $L^{\prime}_i \leftarrow L^{\prime}_i \cup \{$an arbitrary element from $[b]\}$.
      \ENDWHILE
      \ENDFOR
      \STATE $\mathscr{S}^{\prime}\leftarrow L^{\prime}_{1} \times L^{\prime}_{2} \times \cdots \times L^{\prime}_{n}$.
  \end{algorithmic}
\end{algorithm}

    It is clear that the algorithm satisfies Property 3 above.

    Now consider the state of Algorithm 1 at line 18. Let $i$ be such that
    $|L^{\prime}_i|=\mathbf{L_{max}}$. Clearly $L^{\prime}_i$ includes the elements in $L_i$ which are at most $\ell$ many.
    Moreover every new element added to $L'_i$ in the loop spanning lines 8-12 covers
    a section of $[b]$ of width $\tau$, and thus $b/\tau=C\kappa\ell$
    elements can be added.
    Thus $\mathbf{L_{max}}\leq \ell + C \kappa \ell$. At the end of the algorithm every $L^{\prime}_i$ contains
    either $1$ element (which is $\{0\}$) or $\mathbf{L_{max}}$ elements. This gives us Property 1. Note that $C\geq 4$ by construction.

    It is easy to verify that it satisfies Property 2 as well (the $\log b$ factor in the runtime is present
    because the algorithm works with $(\log b)$-bit integers).

    Property 1 and the bound $|\mathcal{I}_\mathscr{S}|\leq \kappa$ together give that the
    number of corners in $\mathscr{S}$ is at most $(\ell + C \kappa \ell)^{\kappa}$.  It is easy to see
    from the algorithm that the area covered by each corner in $\mathscr{S}'$ is at most $\frac{b^n}{(C\kappa\ell)^{\kappa}}$
    (again using the bound on $|\mathcal{I}_\mathscr{S}|$).  Therefore any
    $\epsilon$ fraction of the corners in $\mathscr{S}^{\prime}$ cover an area of at most:
    \[ \epsilon (\ell + C \kappa \ell)^{\kappa} \times \frac{b^n}{(C\kappa\ell)^{\kappa}} =
    \epsilon {(1+\frac{1}{C\kappa})}^{\kappa}\times b^n \underbrace{<}_{C\geq 4} e^{1/3}\epsilon b^{n}< 2\epsilon b^{n}.
    \]
    This gives Property 4.
\qed
\end{pf}

The following lemma is easy and useful; similar statements are given in \cite{BK}.
Note that the lemma critically relies on the $b$-literals being basic.
\begin{lemma}\label{lemma3app2} Let $\fisgfunc$ be expressed as an $s$-way \textsc{Majority} of
\textsc{Parity} of basic $b$-literals. Then for each index $1\leq i \leq n$, there are
at most $2s$ $i$-sensitive values with respect to $f$.
\end{lemma}
\begin{pf}
A literal $\ell$ on variable $x_i$ induces two $i$-sensitive values.
The lemma follows directly from our assumption (see Section~\ref{sec:prelim})
that for each variable $x_i$,
each of the $s$ \textsc{Parity} gates has no more than one incoming literal
which depends on $x_i$.
\qed
\end{pf}

Algorithm 2 is our extension of the $\ghs$ algorithm. It essentially
works by repeatedly running $\ghs$ on the target function $f$ but
restricted to a small (relative to $\bbn$) grid. To upper bound the number
of steps in each of these invocations we will be referring to the result of
Theorem~\ref{thm:analogue}. After each execution of $\ghs$, the hypothesis defined
over the grid is extended to $\bbn$ in a natural way and is tested for
$\eps$-accuracy. If $h$ is not $\eps$-accurate, then a point
where $h$ is incorrect is used to identify a new sensitive value and this
value is used to refine the grid for the next iteration.
The bound on the number of sensitive values from Lemma~\ref{lemma3app2}
lets us bound the number of iterations.
Our theorem about Algorithm~2's performance is the following:

\begin{theorem} \label{thmapp2} Let concept class $\cc$ consist of
$s$-way \textsc{Majority} of $r$-way \textsc{Parity} of basic
$b$-literals such that $s=\poly(n\log b)$ and each $f\in C_{n,b}$
has at most $\kappa(n,b)$ non-trivial indices and at most
$\ell(n,b)$ $i$-sensitive values for each $i=1,\ldots, n$. Then
$\cc$ is efficiently learnable if $r=O(\frac{\log (n \log b)}{\log
\log \kappa \ell})$.
\end{theorem}

\begin{algorithm}[t]
    \caption{An improved algorithm for learning \textsc{Majority} of \textsc{Parity} of basic $b$-literals.}
  \begin{algorithmic}[1]
      \STATE $L_1\leftarrow\{ 0 \}, L_2\leftarrow\{ 0 \},\ldots, L_n\leftarrow\{ 0 \}$.
      \LOOP
      \STATE $\mathscr{S}\leftarrow L_{1} \times L_{2} \times \cdots \times L_{n}$.
      \STATE $\mathscr{S}^{\prime}\leftarrow$ the output of refinement algorithm with input $\mathscr{S}$.
      \STATE One can express $\mathscr{S}^{\prime}= L^{\prime}_{1} \times L^{\prime}_{2} \times \cdots \times L^{\prime}_{n}$. If $L_i\neq\{0\}$
      then $L^{\prime}_{i} = \{ x^{i}_{0},  x^{i}_{1}\ldots, x^{i}_{\mathbf{(L_{max}}-1)}\}$. Let $x^{i}_{0}< x^{i}_{1}< \cdots  < x^{i}_{t-1}$ and
      let $\tau_i: \mathbb{Z}_{\mathbf{L_{max}}} \rightarrow L^{\prime}_{i}$ be the translation function such that $\tau_i(j)=x^{i}_{j}$.
      If $L_i=L^{\prime}_{i}=\{0\}$ then $\tau_i$ is the function simply mapping $0$ to $0$.
      \STATE Invoke $\ghs$ over $f|_{\mathscr{S}^{\prime}}$ with accuracy $\epsilon/8$. This is done
      by simulating $\mathsf{MEM}(f|_{\mathscr{S}^{\prime}}(x_1,\ldots,x_n))$ with $\mathsf{MEM}(f(\tau_1(x_1),\tau_2(x_2),\ldots,\tau_n(x_n)))$.
      Let the output of the algorithm be $g$.
      \STATE Let $h$ be a hypothesis function over $\bbn$ such that $h(x_1,\ldots,x_n)=g(\tau^{-1}_{1}(\lfloor x_1 \rfloor), \ldots,\tau^{-1}_{n}( \lfloor x_n \rfloor))$ ($\lfloor x_i\rfloor$ denotes largest value in $L^{\prime}_i$ less than or equal to $x_i$).
      \IF{$h$ $\epsilon$-approximates $f$}
      \STATE Output $h$ and terminate.
      \ENDIF
      \STATE Perform random membership queries until an element $(x_1,\ldots,x_n)\in \bbn$ is found such that
      $f(\lfloor x_1 \rfloor,\ldots,\lfloor x_n \rfloor)\neq f(x_1,\ldots,x_n)$.
      \STATE Find an index $1\leq i \leq n$ such that $$f(\lfloor x_1 \rfloor,\ldots,\lfloor x_{i-1} \rfloor,x_i, \ldots, x_n)\neq
      f(\lfloor x_1 \rfloor,\ldots,\lfloor x_{i-1} \rfloor,\lfloor x_i \rfloor ,x_{i+1}, \ldots, x_n).$$
      This requires $O(\log n)$ membership queries using binary search.
      \STATE Find a value $\sigma$ such that $\lfloor x_i \rfloor +1\leq \sigma \leq x_i$ and \small
      $$f(\lfloor x_1 \rfloor,\ldots,\lfloor x_{i-1} \rfloor,\sigma-1,x_{i+1}, \ldots, x_n)\neq
      f(\lfloor x_1 \rfloor,\ldots,\lfloor x_{i-1} \rfloor,\sigma ,x_{i+1}, \ldots, x_n).$$\normalsize
      This requires $O(\log b)$ membership queries using binary search.
      \STATE $L_i \leftarrow  L_i \cup \{ \sigma \}$.
      \ENDLOOP
  \end{algorithmic}
\end{algorithm}

\begin{pf}
We assume $b=\omega(\kappa \ell)$ without loss of generality. Otherwise one immediately obtains the result with a direct application of $\ghs$ through Theorem~\ref{thm:analogue}.

We clearly have $\kappa \leq n$ and $\ell \leq 2s.$ By Lemma~\ref{lemma3app2}
there are at most $\kappa\ell = O(ns)$ sensitive values. We will show that
the algorithm finds a new sensitive value at each iteration and terminates before
all sensitive values are found. Therefore the number of iterations will be upper
bounded by $O(ns)$. We will also show that each iteration runs in $\poly(n,\log b,\epsilon^{-1})$
steps. This will establish the desired result.

Let us first establish that step 6 takes at most $\poly(n,\log
b,\epsilon^{-1})$ steps. To observe this it is sufficient to combine
the following facts:
\begin{itemize}
    \item Due to the construction of Algorithm 1 for every non-trivial index $i$ of $f$, $L^{\prime}_{i}$ has
        fixed cardinality $=\mathbf{L_{max}}$. Therefore $\ghs$ could be invoked over the restriction of $f$ onto the grid,
        $f |_{\mathscr{S}^{\prime}}$, without any trouble.
    \item If $f$ is $s$-way \textsc{Majority} of $r$-way \textsc{Parity} of basic $b$-literals, then the function
        obtained by restricting it onto the grid: $f |_{\mathscr{S}^{\prime}}$ could be expressed as
        $t$-way \textsc{Majority} of $u$-way \textsc{Parity} of basic $L$-literals where $t\leq s$, $u\leq r$ and
        $L\leq O(\kappa \ell)$ (due to the $1^{\mathrm{st}}$ property of the refinement).
    \item Due to Theorem~\ref{thm:analogue}, running $\ghs$ over a grid with alphabet size $O(\kappa \ell)$
        in each non-trivial index takes $\poly(n,\log b,\epsilon^{-1})$ time if the dimension of the rectangles are
        $r=O(\frac{\log (n \log b)}{\log \log \kappa \ell})$.
        The key idea here is that running $\ghs$ over this $\kappa \ell$-size alphabet
        lets us replace the ``$b$'' in Theorem~\ref{thm:analogue} with ``$\kappa \ell$''.
\end{itemize}

To check whether if $h$ $\epsilon$-approximates $f$ at step $8$, we may draw
$O(1/\epsilon)\cdot \log(1/\delta)$ uniform random examples and use the membership
oracle to empirically estimate $h$'s accuracy on these examples.  Standard bounds on
sampling show that if the true error rate of $h$ is less than (say) $\eps/2$, then
the empirical error rate on such a sample will be less than $\eps$ with probability
$1 - \delta$. Observe that if all the sensitive values are recovered by the algorithm,
$h$ will $\epsilon$-approximate $f$ with high probability. Indeed, since $g$ $(\epsilon/8)$-approximates
$f |_{\mathscr{S}^{\prime}}$, Property 4 of the refinement guarantees that misclassifying the
function at $\epsilon/8$ fraction of the corners could at most incur an overall error of
$2\epsilon/8=\epsilon/4$. This is because when all the sensitive elements are recovered, for
every corner in $\mathscr{S}^{\prime}$, $h$ either agrees with $f$ or disagrees with $f$ in the entire
region covered by that corner. Thus $h$ will be an $\epsilon/4$ approximator to $f$ with high
probability. This establishes that the algorithm must terminate within $O(ns)$ iterations
of the outer loop.

Locating another sensitive value occurs at steps $11$, $12$ and $13$. Note that $h$ is not an
$\epsilon$-approximator to $f$ because the algorithm moved beyond step $8$. Even if we were to
correct all the mistakes in $g$ this would alter at most $\epsilon/8$ fraction of the
corners in the grid $\mathscr{S}^{\prime}$ and therefore $\epsilon/4$ fraction of the values in
$h$ -- again due to the $4^{\mathrm{th}}$ property of the refinement and the way $h$ is generated.
Therefore for at least  $3\epsilon/4$ fraction of the domain we ought to have
$f(\lfloor x_1 \rfloor,\ldots,\lfloor x_n \rfloor)\neq f(x_1,\ldots,x_n)$ where
$\lfloor x_i\rfloor$ denotes largest value in $L^{\prime}_i$ less than or equal to $x_i$.
Thus the algorithm requires at most $O(1/\epsilon)$ random queries to find such an input in step $11$.

Thus we have observed that steps 6, 8, 11, 12, 13 take at most $\poly(n,\log b,\epsilon^{-1})$ steps.
Therefore each iteration of Algorithm 2 runs in $\poly(n,\log b,\epsilon^{-1})$ steps as claimed.

We note that we have been somewhat cavalier in our treatment of the
failure probabilities for various events.  These include the
possibility of getting an inaccurate estimate of $h$'s error rate in
step 9, or not finding a suitable element $(x_1,\dots,x_n)$ soon
enough in step~11, or having the $\ghs$ algorithm fail to return a
good hypothesis in one of its executions. A standard analysis shows
that all these failure probabilities can be made suitably small so
that the overall failure probability is at most $\delta$ within the
claimed runtime. \qed
\end{pf}

\section{Applications to learning unions of rectangles}

In this section we apply the results we have obtained in
Sections~\ref{sec:firstapproach} and \ref{sec:grid} to obtain results
on learning unions of rectangles and related classes.

\subsection{Learning majorities and unions of many low-dimensional
rectangles} \label{sec:subsec1}

The following lemma will let us apply our algorithm for learning
\textsc{Majority} of \textsc{Parity} of $b$-literals to learn
\textsc{Majority} of \textsc{And} of $b$-literals:

\begin{lemma} \label{lem:majand}
Let $\fisafunc$ be expressible as an $s$-way \textsc{Majority} of
$r$-way \textsc{And} of Boolean literals.  Then $f$ is also
expressible as a $O(n s^2)$-way \textsc{Majority} of $r$-way
\textsc{Parity} of Boolean literals.
\end{lemma}
We note that Krause and Pudl{\'{a}}k also gave a related but
slightly weaker bound in \cite{KP}; they used a probabilistic
argument to show that any $s$-way \textsc{Majority} of \textsc{And}
of Boolean literals can be expressed as an $O(n^2 s^4)$-way
\textsc{Majority} of \textsc{Parity}.  Our boosting-based argument
below closely follows that of \cite[Corollary 13]{JACKSON}.

\begin{pf*}{PROOF OF LEMMA~\ref{lem:majand}.}
Let $f$ be the \textsc{Majority} of $h_1,\ldots,h_s$ where each $h_i$ is an
\textsc{And} gate of fan-in $r$.
By Lemma~\ref{disclemma}, given any distribution $\dist$ there
is some \textsc{And} function $h_j$ such that $|\E_{\dist}[f h_j]|\geq 1/s$.
Moreover the $L_1$-norm of any \textsc{And} function
is at most $3$. To see this observe that one can express \textsc{And} as follows:
\begin{align*}
    \textsc{And}(x_1,\ldots,x_r)&=2 \left(\prod_{i=1}^{r} (\frac{1-x_i}{2}) \right)  -1 = 2 \left(\sum_{S\subseteq\{1,\ldots,r\}} \frac{(-1)^{|S|}}{2^{r}} \chi_S\right) -1 \\
    &=-1 + \frac{2}{2^{r}} + \sum_{|S|\geq 1} \frac{2 (-1)^{|S|}}{2^{r}} \chi_{S}.
\end{align*}
Consequently $L_1($\textsc{And}$_r) \leq 1 + (2^{r})\cdot
\frac{1}{2^{r-1}} =3$ and thus we have $L_1(h_j) \leq  3$.

Now Observation~\ref{simplefact} implies that there must be some
parity function $\chi_a$ such that $|\E_{\dist}[f
\conju{\chi_a}]|\geq 1/4s$, where the variables in $\chi_a$ are a
subset of the variables in $h_j$ -- and thus $\chi_a$ is a parity of
at most $r$ literals.  As in the proof of \cite[Corollary
13]{JACKSON}, we can now apply the boosting algorithm of
\cite{Freund:95}; this algorithm runs for $O(\log(1/\eps)/\gamma^2)$
stages to construct an $\eps$-accurate final hypothesis if it is
given a weak hypothesis with advantage $\gamma$ at each stage. We
choose the weak hypothesis to be a \textsc{Parity} with fan-in at
most $r$ at each stage of boosting, and the above arguments ensure
that each weak hypothesis has advantage at least $1/4s$ at every
stage of boosting. If we boost to accuracy $\eps = {\frac 1 {2^n +
1}}$, then the resulting final hypothesis will have zero error with
respect to $f$ and will be a \textsc{Majority} of
$O(\log(1/\eps)/s^2) = O(ns^2)$ many $r$-way \textsc{Parity}
functions. \qed
\end{pf*}

Note that while this argument does not lead to a computationally
efficient construction of the desired \textsc{Majority} of $r$-way
\textsc{Parity}, it does establish its existence, which is all we
need.

Also note that any union (\textsc{Or}) of $s$ many $r$-dimensional
rectangles can be expressed as an $O(s)$-way \textsc{Majority} of
$r$-dimensional rectangles as well.

Theorem~\ref{thmapp1} and Lemma~\ref{lem:majand} together give us
Theorem~\ref{finthm1}.  (In fact, these results give us learnability
of $s$-way \textsc{Majority} of $r$-way \textsc{And} of $b$-literals
which need not necessarily be basic.)

\subsection{Learning unions of fewer rectangles of higher dimension}
\label{sec:subsec2}

We now show that the number of rectangles $s$ and the dimension bound $r$ of
each rectangle can be traded off against each other in Theorem~\ref{finthm1}
to a limited extent.
We state the results below for the case
$s= \poly(\log(n \log b))$, but one could obtain analogous results
for a range of different choices of $s$.

We require the following lemma:
\begin{lemma}  \label{lem:ks}
Any $s$-term $r$-$\dnf$ can be expressed as an $r^{O(\sqrt{r}\log
s)}$-way \textsc{Majority} of $O(\sqrt{r}\log s)$-way
\textsc{Parity} of Boolean literals.
\end{lemma}
\begin{pf}
\cite[Corollary 13]{KS} states that any $s$-term $r$-$\dnf$ can be
expressed as an $r^{O(\sqrt{r}\log s)}$-way \textsc{Majority} of
$O(\sqrt{r}\log s)$-way \textsc{And}s.  Now recall that the Fourier
representation of an \textsc{And} of $t$ variables is a linear
combination of $2^{t}$ \textsc{Parity}s (or negated \textsc{Parity}s), each
with a coefficient of $1/2^t$ (this Fourier representation is given
explicitly in the proof of Lemma~\ref{lem:majand}).  Clearing this
common denominator, we may simply replace each \textsc{And} that is input the
\textsc{Majority} with the corresponding sum of $2^t$
\textsc{Parity}s (or negated \textsc{Parity}s). This gives the
lemma. \qed
\end{pf}

Now we can prove Theorem~\ref{finthm2}, which gives
us roughly a quadratic improvement in the dimension
$r$ of rectangles over Theorem~\ref{finthm1} when $s=\poly(\log(n \log b))$.

\begin{pf*}{PROOF OF THEOREM~\ref{finthm2}.}
First note that by Lemma~\ref{lemma3app2}, any function in $C_{n,b}$
(as defined by Section~\ref{sec:lrnmodel}) can have at most
$\kappa=O(rs)=\poly (\log (n \log b))$ non-trivial indices, and at
most $\ell=O(s)=\poly (\log (n \log b))$ many $i$-sensitive values
for all $i=1,\ldots,n$. Now use Lemma~\ref{lem:ks} to express any
function in $C_{n,b}$ as an $s^{\prime}$-way \textsc{Majority} of
$r^{\prime}$-way \textsc{Parity} of basic $b$-literals where
$s^{\prime}= r^{O(\sqrt{r}\log s)}=\poly(n \log b)$ and
$r^{\prime}=O(\sqrt{r}\log s)=O(\frac{\log (n \log b)} {\log \log
\log (n \log b)})$. Finally, apply Theorem~\ref{thmapp2} to obtain
the desired result. \qed
\end{pf*}

Note that it is possible to obtain a similar result for learning
$\poly(\log(n \log b))$-way  union of $O(\frac{\log^2 (n \log b)}
{(\log \log(n \log b))^4})$-way \textsc{And} of $b$-literals if one
were to invoke Theorem~\ref{thmapp1}.

\subsection{Learning majorities of unions of disjoint rectangles}
\label{sec:subsec3}

A set $\{R_1,\dots,R_s\}$ of rectangles is said to be {\em disjoint}
if every input $x \in \bbn$ satisfies at most one of the rectangles.
Learning unions of disjoint rectangles over $\bbn$  was
studied by \cite{BK}, and is a natural
analogue over $\bbn$ of learning ``disjoint $\dnf$'' which has been
well studied in the Boolean domain (see e.g. \cite{KHARDON,ABKKPR}).

We observe that when disjoint rectangles are considered
Theorem~\ref{finthm1} extends to the concept class of majority of
unions of disjoint rectangles. This extension relies on the easily
verified fact that if $f_1,\dots,f_t$ are functions from $\bbn$ to
$\bn$ such that each $x$ satisfies at most one $f_i$, then the
function \textsc{Or}$(f_1,\dots,f_t)$ satisfies $
L_1(\textsc{Or}(f_1,\dots,f_t)) = O(L_1(f_1) + \cdots + L_1(f_t)).
$ This fact lets us apply the argument behind Theorem~\ref{mthm1}
without modification, and we obtain Corollary~\ref{fincor2}. Note
that only the rectangles connected to the same \textsc{Or} gate must
be disjoint in order to invoke Corollary~\ref{fincor2}.

\section{Conclusions and future work}

For future work, besides the obvious goals of strengthening our positive
results, we feel that it would be interesting to explore the limitations
of current techniques for learning unions of rectangles over $\bbn$.
At this point we cannot rule out the possibility that the Generalized Harmonic
Sieve algorithm is in fact a poly$(n,s,\log b)$-time algorithm for learning
unions of $s$ arbitrary rectangles over $\bbn$.  Can evidence for
or against this possibility be given?  For example, can one
show that the representational power of the hypotheses
which the Generalized Harmonic Sieve algorithm produces
(when run for $\poly(n,s,\log b)$ many stages) is -- or is not -- sufficient
to express high-accuracy approximators to arbitrary
unions of $s$ rectangles over $\bbn$?

\section{Acknowledgement}

We thank the anonymous journal referees whose helpful suggestions
improved the presentation of this paper.

\end{document}